\documentclass[letterpaper, 10 pt, conference]{ieeeconf}  %

\IEEEoverridecommandlockouts                              %

\usepackage[colorinlistoftodos,prependcaption,textsize=tiny]{todonotes}
\usepackage{graphicx} %
\usepackage{caption}
\usepackage{subcaption}
\usepackage{booktabs}
\usepackage{makecell}
\usepackage{gensymb}
\usepackage{flushend}
\usepackage[hidelinks]{hyperref}
\newcommand{\ts}{\textsuperscript}

\usepackage[top=54pt, left=54pt, right=54pt, bottom=57pt]{geometry}

\title{\LARGE \bf
Reactive Neural Path Planning with Dynamic Obstacle Avoidance \\ in a Condensed Configuration Space

}

\author{Lea Steffen$^{1}$, Tobias Weyer$^{1}$, Stefan Ulbrich$^{1}$, Arne Roennau$^{1}$and R\"udiger Dillmann$^{1}$%

\thanks{$^{1}$All authors are with FZI Research Center for Information Technology,
        76131 Karlsruhe, Germany
        {\tt\small steffen@fzi.de}}%
}

\begin{document}

\maketitle
\thispagestyle{empty}
\pagestyle{empty}

\begin{abstract}
We present a biologically inspired approach for path planning with dynamic obstacle avoidance. Path planning is performed in a condensed configuration space of a robot generated by self-organizing neural networks (SONN).
The robot itself and static as well as dynamic obstacles are mapped from the Cartesian task space into the configuration space by precomputed kinematics.
The condensed space represents a cognitive map of the environment, which is inspired by place cells and the concept of cognitive maps in mammalian brains. 
The generation of training data as well as the evaluation were performed on a real industrial robot accompanied by simulations. To evaluate the reactive collision-free online planning within a changing environment, a demonstrator was realized. Then, a comparative study regarding sample-based planners was carried out. 
So we could show that the robot is able to operate in dynamically changing environments and re-plan its motion trajectories within impressing 0.02 seconds, which proofs the real-time capability of our concept.
\end{abstract}

\section{INTRODUCTION} \label{sec:introduction}
Up until now, industrial robots have been predominantly used in cages with large safety areas. However, the role of human-robot interaction constantly and drastically increases, from automotive assembly to electronic component manufacturing. Thus, the need to integrate robots into dynamic environments becomes evident.
Cooperative tasks require the skill to perform a goal directed motion while taking into account a dynamic environment. While humans master this problem with ease, reactive and collision-free motion planning is extremely challenging for robots. 
Conventional motion planning for robots can be divided in two principal approaches:
First, planning of end-effector trajectories in the task space and subsequent calculation of resulting joint angles by means of inverse kinematic (IK). 
However, this faces redundancies and may lead to unpleasant jumps in the joint angles. 
Second, the more elegant variant is the direct path planning in the N-dimensional configuration space (C-space), where N is the degree of freedom (DOF). Hereby, the robot is represented as a point and the motion planning becomes a pure path planning problem, which has great advantages, especially for obstacle avoidance.
However, obstacles have to be projected to the C-space, by blocking respective configurations, and the dimensionality of the search space directly dependents on a robot's number of DOF. Thus, the complexity of path planning in the C-space significantly increases for robots with many DOFs and cluttered environments.
In this case, traditional path planning algorithms are very inefficient as they suffer the curse of dimensionality. Particularly, complete and optimal search algorithms such as Wavefront, Dijkstra’s or A* struggle to provide solutions for more than 3D within a reasonable amount of computation time. Sample-based planners try to reduce the computational burden using random samples in the C-space and have become the mainstream method for robots with many DOF~\cite{Liu2021}.
However, their computational costs are still exponentially increasing with the number of DOF and especially when it comes to dynamically changing environments or narrow passages, the conventional sample-based planners have difficulties in online applications~\cite{Kiesel2017, Hsu2003, Szkandera2020, Orthey2021}.
Mapping obstacles from task to C-space is far from trivial. The amount of memory needed for a complete C-space map increases exponentially with the number of DOF \cite{Wise2001}. %
This is true for geometric methods, which try to capture the shape of obstacles, reduce it to basic geometrical features and transform them to the C-space~\cite{Brost1989, Varadhan2006}. %
\begin{figure}[h]
	\centering
	\includegraphics[width=0.9\linewidth]{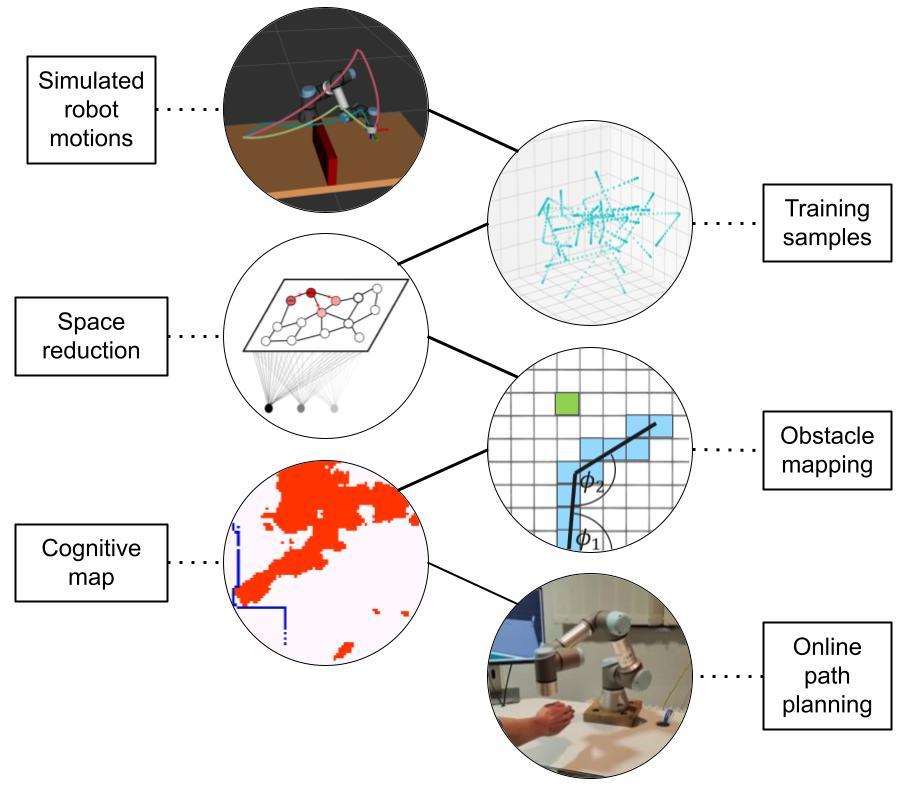}  
	\caption{Pick-and-place movements are used as training data. These sample trajectories are fed to a SONN. Its output space functions as a reduced search space which allows the use of look up tables for collision detection. The generated cognitive map enables fast and flexible path planning.}
	\label{fig:planner_pipeline}
\end{figure}
An analytical method is presented in~\cite{DAvella2003}, whereby parametric equations for the links of a robot are developed by means of its kinematic motion constraints. As the equation set must be checked for each obstacle, this accumulates quickly and prevents on-line planning for new obstacles. 
A technique combining support vector machines and geometric approximation is presented in~\cite{Pan2015}. The authors claim to achieve real-time capability due to massive GPU-parallelization.
In~\cite{Ward2007}, a method is proposed to determine guaranteed collision-free areas of the C-space. It is quite sophisticated regarding performance and memory, but it does not capture all collision-free configurations, which makes it impractical for cluttered environments.
Sample-based motion planners are also build upon the concept of capturing the free C-space.
In~\cite{Han2021} the C-space is sampled via an RRT. The samples are validated to be collision-free to build a free C-space approximation. Besides common issues with narrow passages, random sampling in a high-dimensional search space is computationally quite inefficient. 
Another fast approach for projecting obstacles from Cartesian to C-space is a bidirectional lookup table~\cite{Wu2005, Xie2020, Huerta-Chua2021}. %
This method allows a very fast mapping from the task to the C-space, however, the amount of memory increases exponentially with the DOF of the robot and especially for many DOF the discretization must be very sparse to allow the storage of the whole C-space. 
In this paper we apply a bidirectional lookup table and address the issue of an expansive configuration space with neural self-organization as shown in~\autoref{fig:planner_pipeline}.
\\
The remainder of this paper is structured as follows: \autoref{sec:system_architecture} presents the general idea how to reduce a C-space by means of self-organization and subsequently use it for path planning. \autoref{sec:cognitive_map}, describes the context of the biological model and our implementation of it. In \autoref{sec:obstalce_transform} it is depicted how obstacles are mapped from the task to the C-space. The evaluation in \autoref{sec:results} begins with the robot demonstrator and the visualization of the output space. This is complemented with a comparison of the Wavefront and Dijkstra's algorithm executed in our reduced space. Subsequently we compare two Self-organizing neural networks (SONN) for obstacle avoidance and finally evaluate our approach against modern sample-based planners. We provide an overall conclusion in \autoref{sec:conclusions}.

\section{REDUCED CONFIGURATION SPACE} \label{sec:system_architecture} 
The proposed approach is inspired by the self-organizing topological structures in the brain and the concept of cognitive maps and place cells. A SONN is trained to build a cognitive map in the C-space of a robot while reducing the complexity of this n-dimensional C-space.
SONNs are an unsupervised learning technique often used for clustering, classification or, as in our case, dimension reduction of an input space~\cite{Miljkovic2017, VanHulle2012}.
Its special feature is that they preserve the topology of the input data during learning. Hence, neighboring points in the output space are also neighboring in the input space.
SONNs are a useful tool for configuration space dimension reduction while preserving its underlying topology~\cite{Steffen2021_dimreduction, Steffen2022_dimreduction2}. 
A proof-of-concept is given in \cite{Steffen2021_dimreduction}, with human motions as input data. The concept was refined in \cite{Steffen2022_dimreduction2} and transferred to a robot's kinematics. Additionally, a comprehensive study regarding the performance of different SONN models was carried out in~\cite{Steffen2022_dimreduction2}. \\
Based on the most performant versions, we now present a biologically inspired collision-free path planning method for robot arms with multiple DOF. It is carried out directly in the C-space of the robot and hence, no calculations of kinematics nor IK are needed during run time. 
\begin{figure*}[h!]
	\centering
	\includegraphics[width=0.9\textwidth]{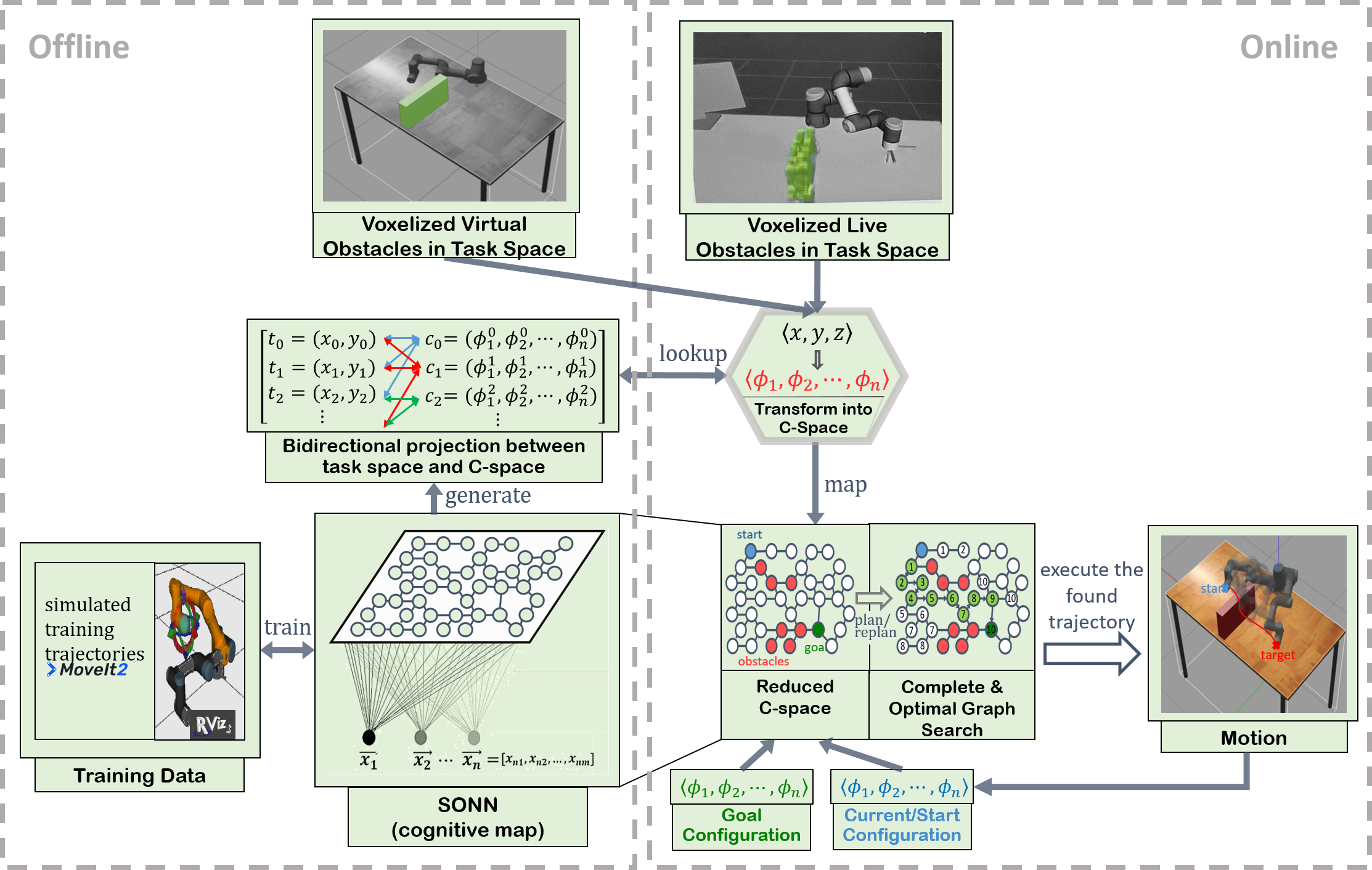}
	\caption{\textit{Bottom left:} Training of the SONN to create a reduced C-space representation. \textit{Top:} Transformation of obstacles into the reduced C-space to allow collision-free path planning using a lookup table. \textit{Bottom right:} Path planning in the reduced free C-space and the execution of the found trajectory with the robot. The SONN training is performed offline while the other 2 components are executed online in real-time.}
	\label{fig:konzeptskizze}
\end{figure*}
In~\autoref{fig:konzeptskizze}, the structure of the proposed system and its three main parts are shown. The heart of the approach is a SONN, which can be seen at the bottom left. It represents a cognitive map to reduce the complexity of the robot C-space, as depicted in~\autoref{sec:cognitive_map}. Above, a bi-directional lookup table is shown, which is created after the SONN is trained. It links the joint states with the complete robot coverage of the task space and incorporates obstacles from the task space as described in~\autoref{sec:obstalce_transform}. At the bottom right it is visualized how the final path planning can take place in the online phase. \\
The SONN is firstly trained with trajectories of a simulated UR3e robot. To take full advantage of the complexity reduction of this approach, only a sub manifold of the robot’s C-space was used as training data. Thus, reaching motions of a simple pick \& place were recorded. A probabilistic sample-based planner, the RRTConnect~\cite{Kuffner2000} %
of the Open Motion Planning Library~\cite{Sucan2012} was used to generate trajectories. 1230 training trajectories were recorded with a total of 50 728 sample points. 
To integrate intuitive motions, manual hand guidance were recorded as additional training data.\\
The choice of the most suited network structure for our specific use case is based on the findings of~\cite{Steffen2022_dimreduction2}. Nine models were tested, which are all based either on the well known Self-Organizing Map (SOM)~\cite{Kohonen1982} or the Growing Neural Gas (GNG)~\cite{Fritzke1995}. 
In the output space of the SOM, neurons are organized in a 2D grid and each neuron has a weight vector. The network size was chosen to 10\hspace{1.5pt}000 neurons.
During the training phase, samples from the training trajectories are presented
individually to the network as input. The distance between the weight vectors of the neurons is compared with the input samples, the most similar one is chosen as the Best Matching Unit (BMU). In the update step, the BMU is adapted towards the input sample and with a lower learning factor also its topological neighbors. During the application phase, the BMU is used to classify the input. The number of neurons must be determined right at the beginning. 
In contrast, GNGs start with very few neurons and the network grows step by step. 
To allow a fair comparison to the SOM-based model, the GNG was grown up to a size of 10\hspace{1.5pt}000 neurons.
In this process, topological connections between the neurons are learned explicitly and always between the BMU and the 2\ts{nd} BMU, that is the most similar and the second most similar neuron to the input sample.
In the GNGs, all neurons connected to the BMU count as neighbor neurons, which are then also matched to the input sample in the update step. After a certain number of learning iterations, another neuron is inserted between the two neurons with the largest error to expand the network where the input space is still poorly mapped. The resulting structure is now very similar to graphs of probabilistic sample-based planning algorithms. With joint angle configurations as input, the neurons correspond to samples in the C-space.
However, unlike the probabilistic sample-based planners, the samples are not randomly distributed in space but the distribution of the training data is learned along with the topological connections. Thus, the connections represent the topology.
The final path planning takes place in the online phase. Therefore, BMUs of the current robot joint angle configuration and the BMU of any target configuration are determined as start and target nodes.
It was found in~\cite{Steffen2021_dimreduction} that path planning in a reduced C-space generally works for all SONN models, but with different quality of the generated motion. 
The spatial coverage of SOM-based versions is significantly worse than for GNG-based models. SOMs fold in space due to their rigid structure and therefore cannot map the trajectories topologically in an optimal way. Hence, their generated paths show loops and detours, while most GNG-based versions could actually generate nicely targeted motions.
Furthermore, it was found that the GNG itself performed better than its tested extensions.
However, due to their rigid 2D structure, the SOM models are very suitable to visualize the proposed path planning approach.
Also, the capabilities of the networks in terms of obstacle avoidance have not yet been considered in previous work~\cite{Steffen2022_dimreduction2}. We expect the GNG to perform better here as well, but will evaluate its performance against the most promising SOM candidate, the $\gamma$-SOM \cite{Estevez2009}, in \autoref{sec:som_vs_gng}. The $\gamma$-SOM is designed for temporal sequence processing, as it includes an adjustable memory structure \cite{Estevez2009}. Parameters to train the SONNs were taken from \cite{Steffen2022_dimreduction2}.

\section{Cognitive Map} \label{sec:cognitive_map}
In 1948, Tolman introduced the concept of cognitive maps for mammalians to navigate through their spatial environment \cite{Tolman1948}. 
A cognitive map is the mental representation of a geographical space, or rather of the relationships of locations in the environment. 
The model of cognitive maps of space was substantiated by the discovery of neural place cells in the hippocampus \cite{OKeefe1971, OKeefe1978, Derdikman2010}, whose activity increases if the subject enters a certain location in the environment. Thus, each place cell corresponds to certain locations in the environment. Thereby, the place cells represent the current location as well as past and future locations. This enables not only self-localization but also planning for future positions in space and the path to get there.
This concept of place cells and cognitive maps for navigation of mammalians in the two-dimensional space is transferred to the multi-dimensional C-space of robots. The place cells are represented by the neurons in the output space of the SONN, which creates a cognitive map. %
Each neuron contains a learned joint configuration state that corresponds to a point in the n-dimensional C-space. \\
The $\textrm{BMU}_s$ for the current joint configuration of the robot marks the current position of the robot on the cognitive map, similar to the activation pattern of place cells in the hippocampus if the subject enters a certain location in the environment. 
For path planning, also the $\textrm{BMU}_g$ for the goal configuration is found. Then a graph search algorithm is used to find a path between $\textrm{BMU}_s$ and $\textrm{BMU}_g$ along the topological connections between the neurons. 
To ensure a collision-free path planning, all neurons in the output space of the SONN are blocked for the graph search if their configuration would lead to a collision. 
Since the output space of the SONN represents a reduced space of the full C-space, complete and optimal search algorithms can be used, which would be far too expensive in the full C-space. In~\cite{Steffen2021_dimreduction, Steffen2022_dimreduction2}, the Wavefront algorithm was used as it provided a promising outlook for a very fast and efficient implementation with SNNs on neuromorphic hardware. 
However, a breadth-first search like the Wavefront is ideal for grids and the generated search space of the proposed approach corresponds rather to a weighted graph than to a grid. This is due to the fact that the neurons' learned weights are not evenly distributed in the C-space and therefore show varying distances to their topological neighbored neurons. In contrast to the Wavefront, Dijkstra’s algorithm considers weighted edges between nodes \cite{Huang2009}. Therefore, we included Dijkstra’s algorithm alongside the Wavefront for the evaluation in \autoref{sec:wavefornt_vs_dijkstra}.

\section{Obstacle Avoidance} \label{sec:obstalce_transform}
\begin{figure}[hbt!]
	\centering
	\includegraphics[width=0.45\textwidth]{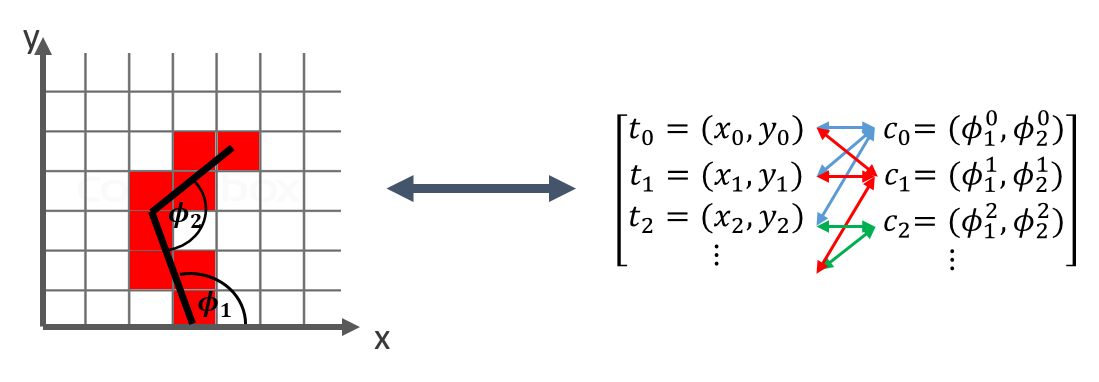}
	\caption{Illustration of a bidirectional lookup table for a two-link robot. The robot is depicted in its discretized 2D work space. The red cells mark the occupied cells of the robot for the joint configuration $c_1=(\phi_1,\phi_2)$. The bidirectional lookup table associates the occupied task space cells $t_i$ with the joint configuration $c_1$. The individual task space cells $t_i$ are not only associated with $c_1$ but also with all other configurations $c_j$. Which would lead to a coverage of the respective task space cell by the robot. If for any task space cell an obstacle is detected, corresponding configurations are blocked for path planning. Inspired by~\cite{Xie2020}.}
	\label{fig:bidirectional_lookup}
\end{figure}

In order to find collision-free paths in the output space of the SONN, neurons which would lead to collisions are blocked for the graph search. Therefore, obstacles in the task space must be mapped into the reduced C-space, to block the respective neurons, as visualized in the top center of \autoref{fig:konzeptskizze}. 
Especially for dynamic obstacles and real-time applications, this mapping must be very fast and highly efficient.
Lookup tables have proven to fulfill these requirements \cite{Wu2005, Xie2020, Huerta-Chua2021}. %
During the generation of the lookup table, the neurons are associated with the corresponding robot task space coverage. 
Therefore, the task space as well as the C-space are discretized into cell grids. 
Then, for every Cartesian point $t_T \in T$ in the task space all C-space points $c_C \in C$ are stored, for which the robot would touch any $t_i$ with any link or joint.
The working principle of the lookup table is exemplary illustrated for a two-link robot in \autoref{fig:bidirectional_lookup}.
The associations between the C-space cells and the task space coverage is calculated via forward kinematics. The stored results enable a very fast and efficient mapping during run time.
The individual robot model parts were inflated before creating the lookup table to establish a safety distance between obstacles and the robot.
The amount of memory increases exponentially with the number of DOF. Hence, for manipulators with many DOF, the discretization must be very sparse to allow the storage of the whole C-space.
However, for the proposed approach only the reduced C-space must be associated with the task space. In other words, the reduced C-space provides a natural discretization of the C-space. Thus, the number of joint configurations associated with the respective task space coverage of the robot diminishes to a reasonable size. 
But, due to the learned representation of the C-space, the sparsely sampled distribution, or rather discretization, ensures a reasonable approximation of important joint states. Every task space cell is associated with several neurons. Thus, if an obstacle is detected covering a task space cell, all neurons with configurations that would lead to a collision are added to a set of blocked neurons.
To avoid collisions with new or dynamic obstacles or to find shorter paths if obstacles vanish, the path is constantly re-planned every 0.3s. To ensure that possibly shorter paths are not the result of short-term voxelization errors, a new path is only accepted if one of 3 criteria applies:
\begin{itemize}
	\item No previous path was found.
	\item The current unsmoothed path would cause a collision as  one of its nodes becomes blocked.
	\item The new unsmoothed path is shorter than the current unsmoothed path and stable for a certain time.
\end{itemize}
If a new path is accepted, it is smoothed. Additionally, it is checked whether the smoothed path is still collision-free, thus, the BMUs of the smoothed configuration states are not blocked. If this is not the case, the smoothing degree is reduced step by step. Subsequently, the smoothed new path is sent to the trajectory controller.

\section{RESULTS AND EXPERIMENTS} \label{sec:results}
\begin{figure*}[t]
	\centering %
	\begin{subfigure}{0.193\textwidth}
		\includegraphics[width=\linewidth]{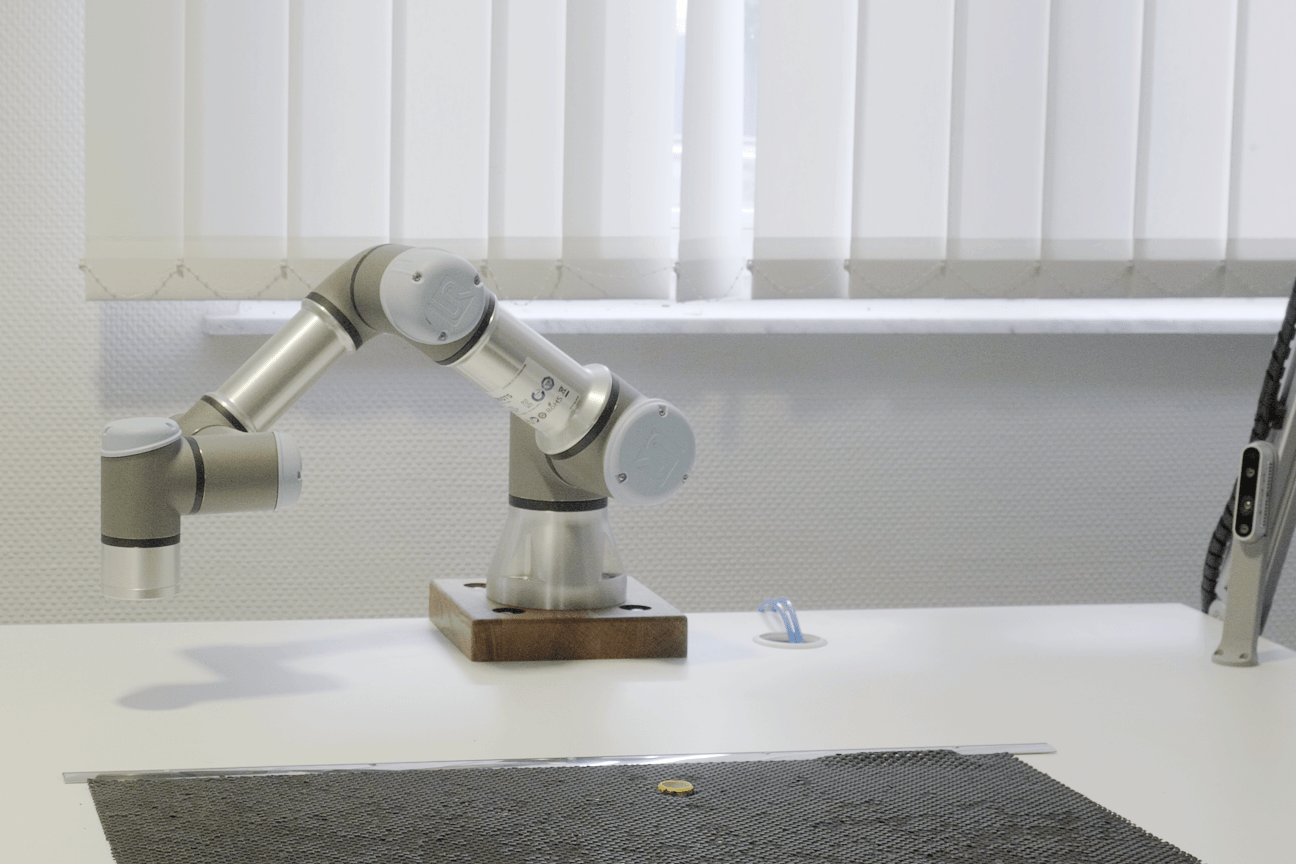}
	\end{subfigure}\hfil %
	\begin{subfigure}{0.193\textwidth}
		\includegraphics[width=\linewidth]{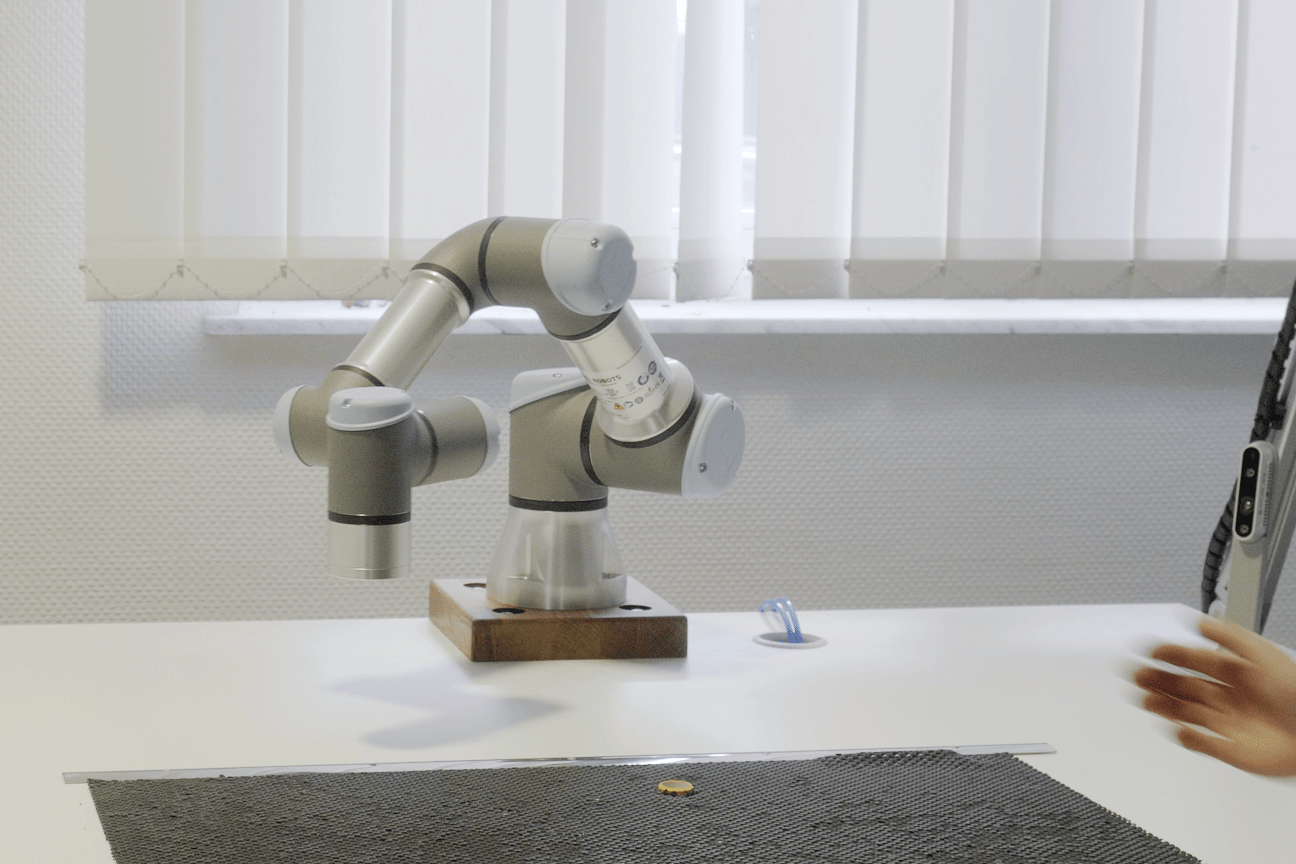}
	\end{subfigure}\hfil %
	\begin{subfigure}{0.193\textwidth}
		\includegraphics[width=\linewidth]{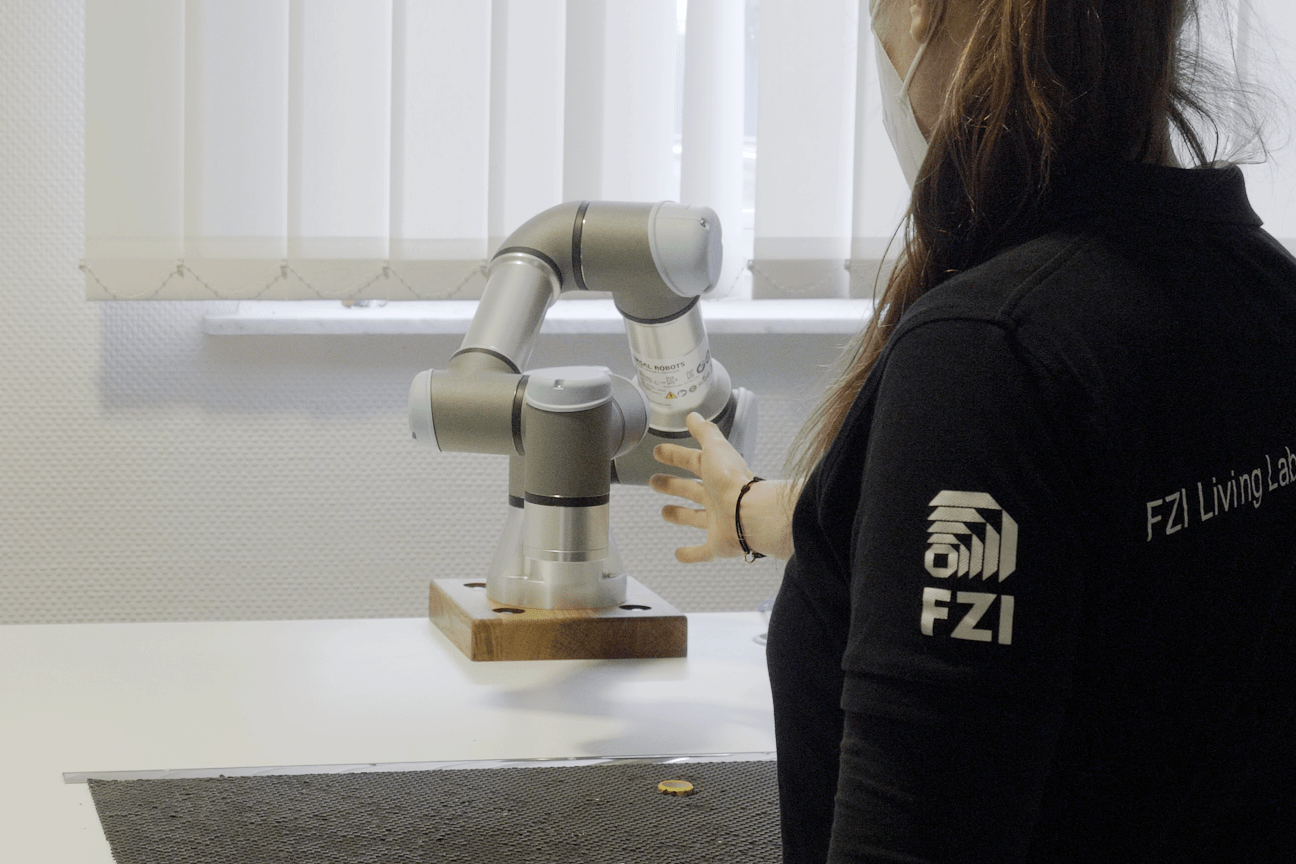}
	\end{subfigure}
	\begin{subfigure}{0.193\textwidth}
		\includegraphics[width=\linewidth]{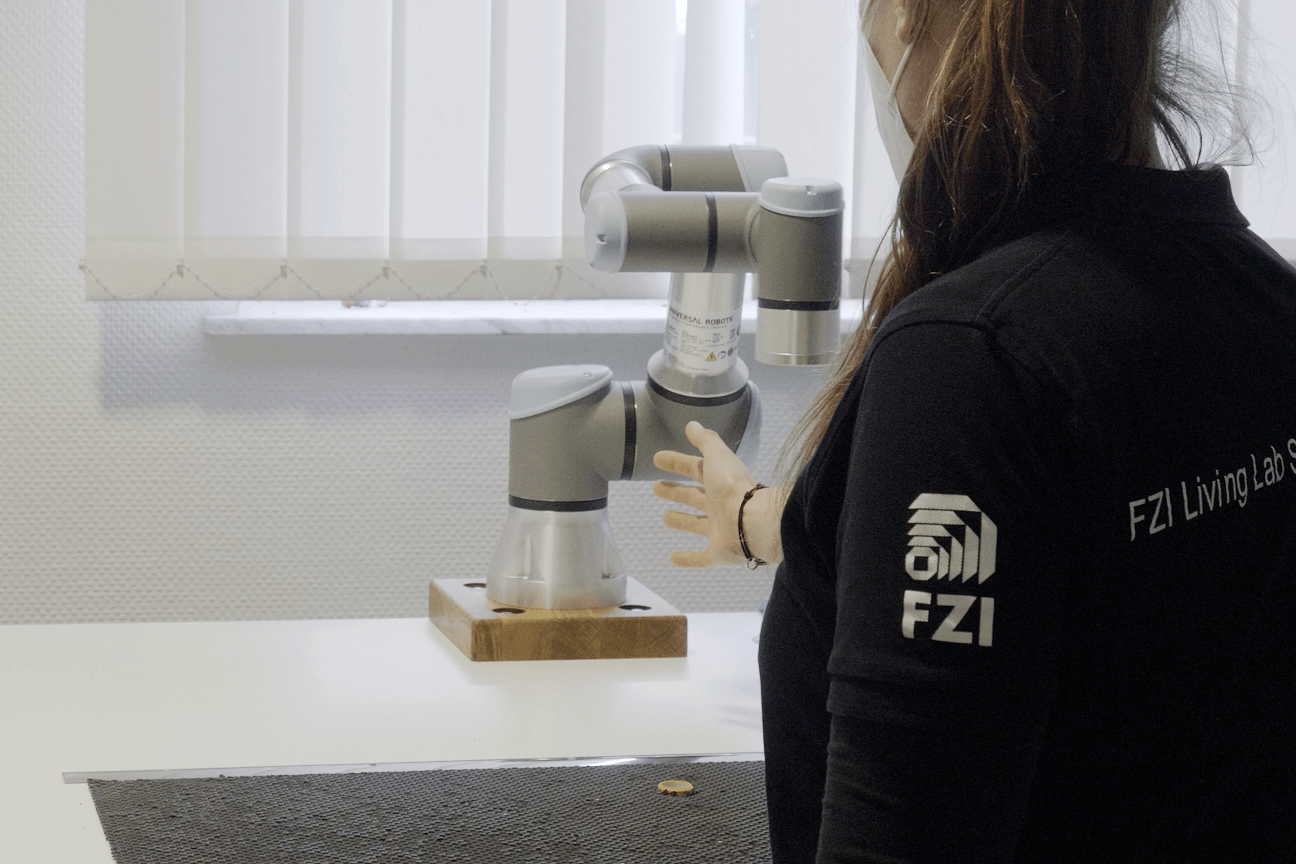}
	\end{subfigure}
	\begin{subfigure}{0.193\textwidth}
		\includegraphics[width=\linewidth]{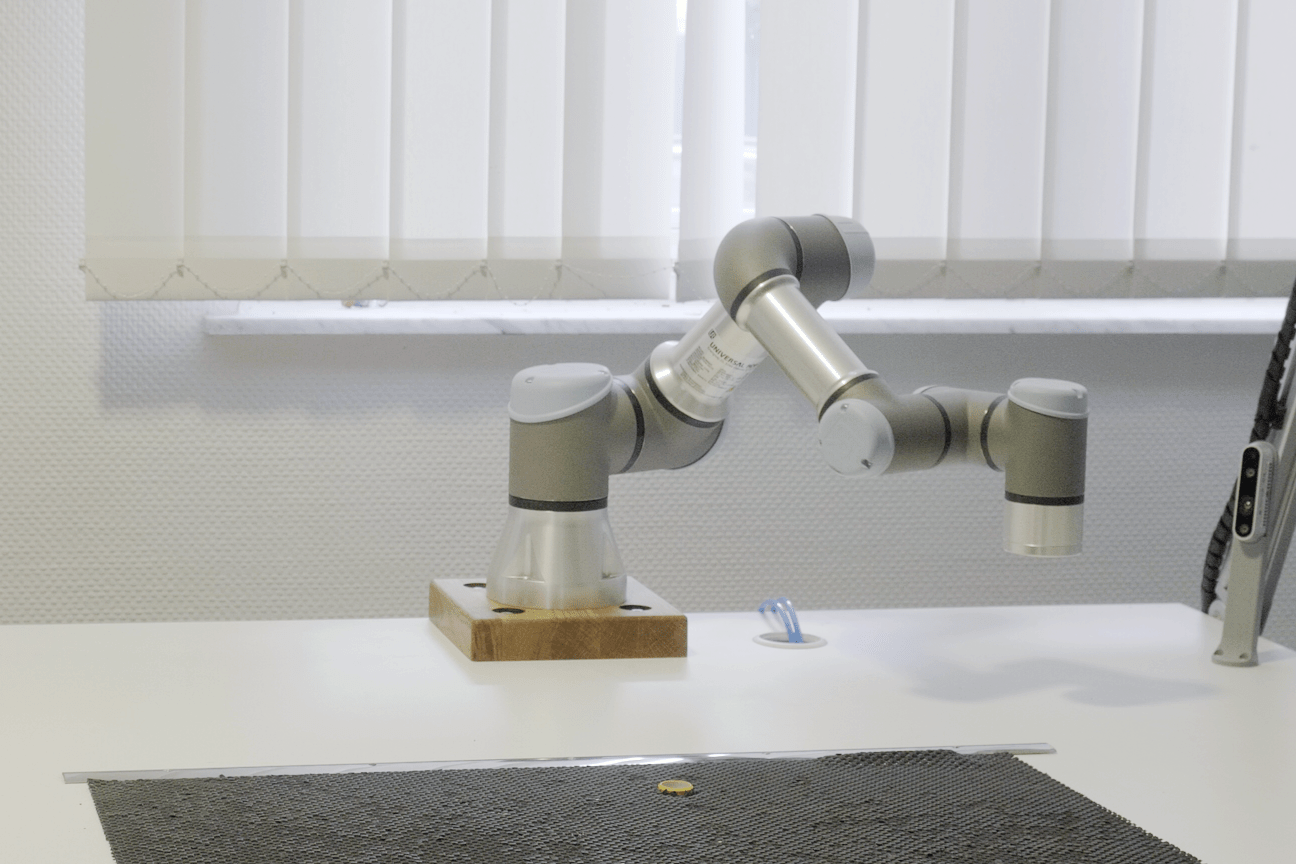}
	\end{subfigure}
	
	\medskip
	\begin{subfigure}{0.193\textwidth}
		\includegraphics[width=\linewidth]{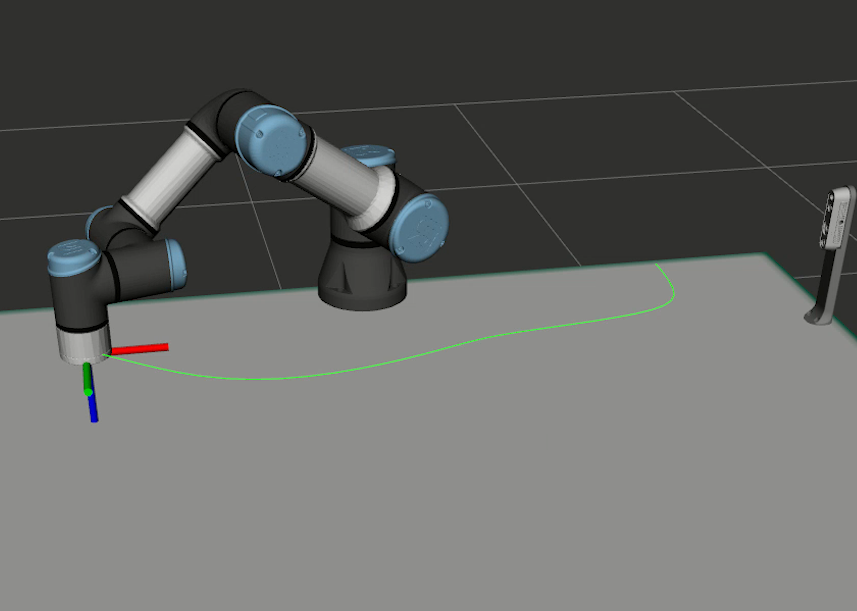}
	\end{subfigure}\hfil %
	\begin{subfigure}{0.193\textwidth}
		\includegraphics[width=\linewidth]{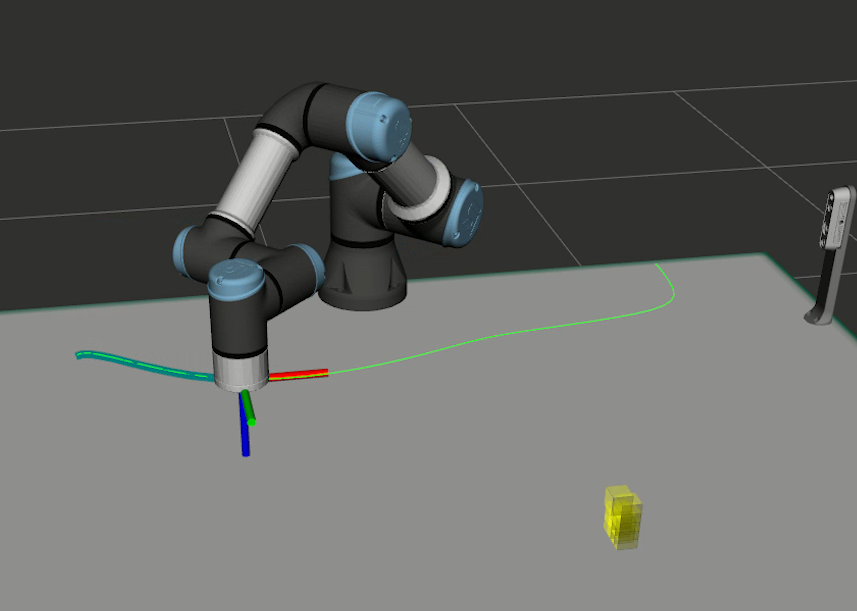}
	\end{subfigure}\hfil %
	\begin{subfigure}{0.193\textwidth}
		\includegraphics[width=\linewidth]{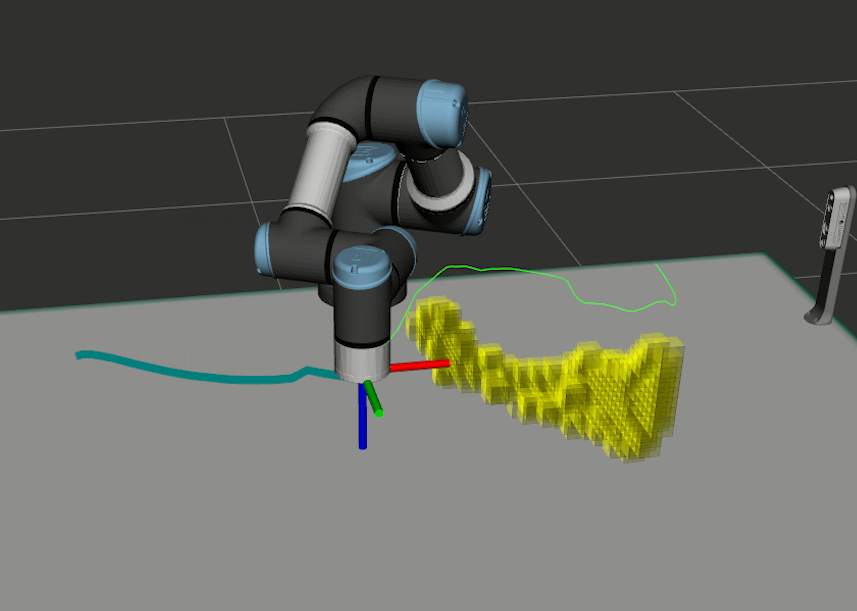}
	\end{subfigure}
	\begin{subfigure}{0.193\textwidth}
		\includegraphics[width=\linewidth]{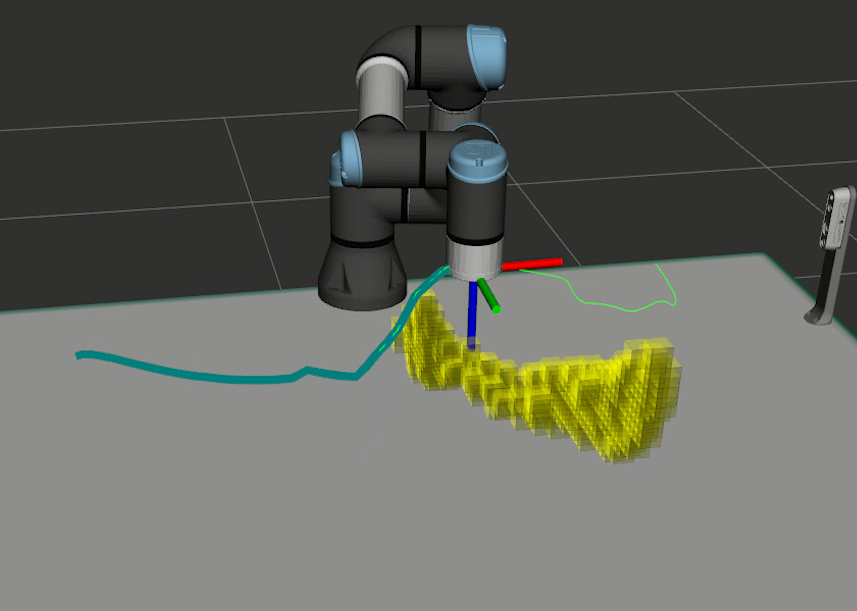}
	\end{subfigure}
	\begin{subfigure}{0.193\textwidth}
		\includegraphics[width=\linewidth]{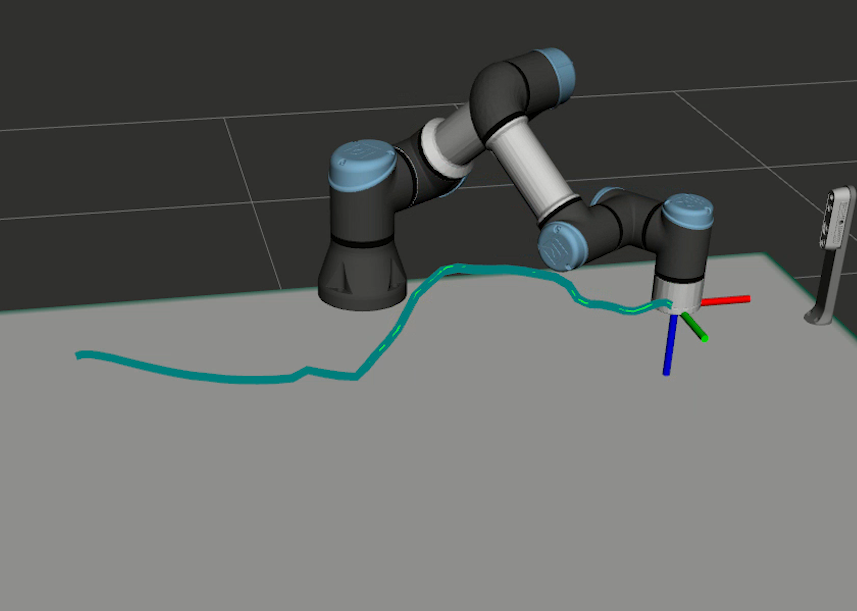}
	\end{subfigure}
	\caption{Photo series of the reactive collision-free motion planning. The first row shows the actual demonstrator and the second one the planned and re-planned path in simulation. The yellow voxels are the detected obstacles by the vision component.
		The green line is the planned and the cyan line the executed end-effector path.}
	\label{fig:series_motion_planning}
\end{figure*}
To evaluate the proposed approach, experiments were performed on a real robot in a work space setup with a vision system including four Intel RealSense D435. The environment detection and voxelization was realized with a software called GPU-Voxels~\cite{Hermann2014}. As the vision component, GPU-Voxels, runs with ROS1~\cite{Quigley2009} and our approach is implemented with ROS2~\cite{Thomas2014}, the ROS1 to ROS2 bridge\footnote{\url{https://index.ros.org/p/ros1_bridge/}} was used.
The planner and the vision component run on separate PCs while the ROS bridge runs on the same PC as the planner and publishes the voxel information to the ROS2 action client.
\autoref{fig:series_motion_planning} shows how a dynamic obstacle is detected and circumvented.
A hand interferes with the initially planned path and the robot re-plans its motion in less than 0.02s and avoids the obstacle. For more details on the demonstrator, see the video supporting this paper. \\
Path planning is performed in the output space of the SONN that is stored in 2D-array structure with a third dimension for the joint configurations. These can be visualized by bitmaps, as shown in~\autoref{fig:sim_and_maps} on the right. 
Every pixel in the bitmap corresponds to one neuron which contains a joint configuration vector. Neurons are marked in red, if they are blocked because their configuration would lead to a collision.
For the two-dimensional structure of the SOM, two adjacent neurons in the bitmap are also topologically connected in the output space of the SONN.
Since there is no rigid structure for the GNG, two neighbored pixel in the bitmap representing neurons are in general not topologically connected but have topological connections to other neurons. 
Thus, for the cognitive maps of the SOMs in~\autoref{fig:gammaSOM_wavefront},~\subref{fig:gammaSOM_dijkstra}~\&~\subref{fig:gammaSOM_dijkstra_smoothed}, coherent obstacles are shown as red areas and a visible path is displayed in blue. In contrast, the activity patterns of the GNGs in \autoref{fig:GNG_dijkstra}~\&~ \subref{fig:GNG_dijkstra_smoothed} are wildly distributed over the whole map and neither obstacles nor the path is a coherent structure. \\
In the following, different aspects of the proposed approach are further analyzed and a comparison with modern sample-based planners is given. To ensure reproducibility the experiments were carried out in a simulated environment. Thereby, only the simulated training data was used for training.

\subsection{Wavefront vs. Dijkstra’s} \label{sec:wavefornt_vs_dijkstra}
\begin{figure}
	\begin{subfigure}{.49\textwidth}
		\centering
		\includegraphics[height=3cm]{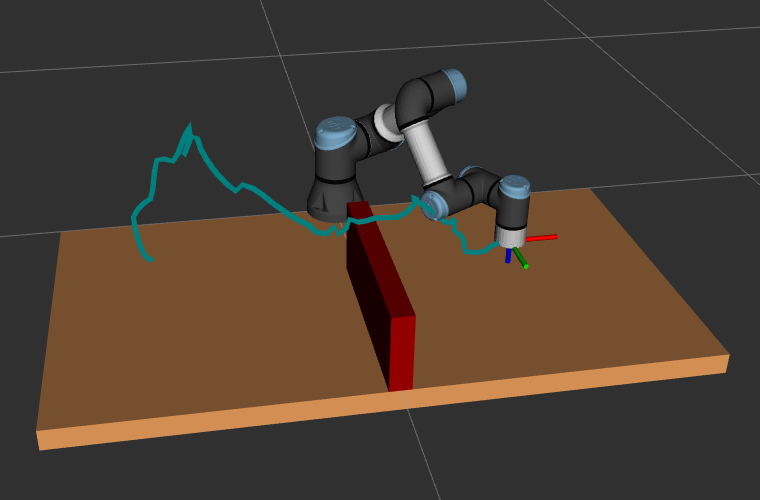}  
        \hspace{0.35cm}
		\includegraphics[height=3cm]{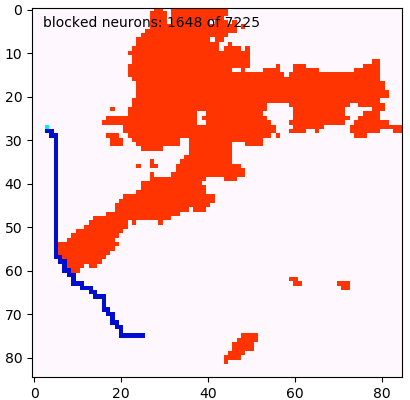}  
		\caption{$\gamma$-SOM, Wavefront path planning}
		\label{fig:gammaSOM_wavefront}
	\end{subfigure}
	\begin{subfigure}{.49\textwidth}
		\centering
		\includegraphics[height=3cm]{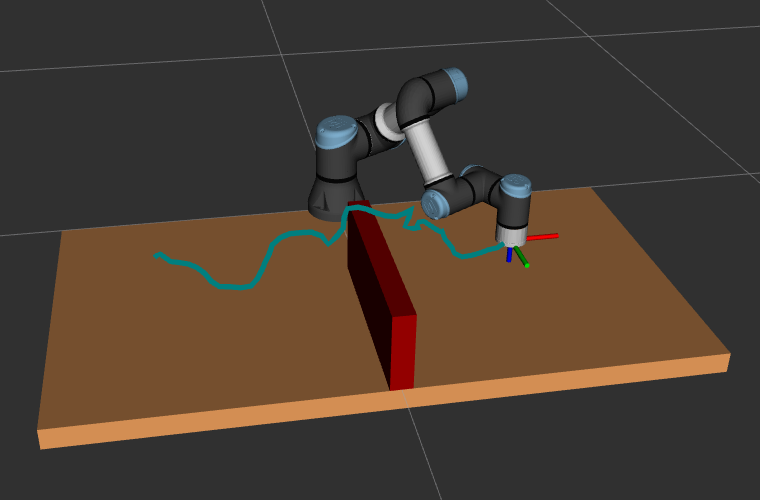}  
        \hspace{0.35cm}
		\includegraphics[height=3cm]{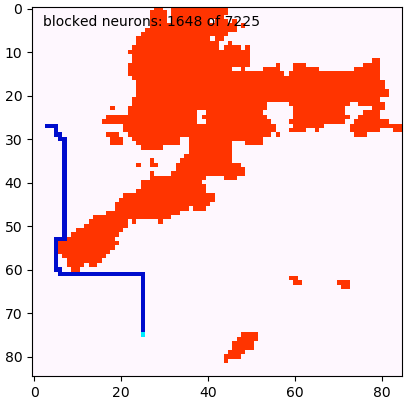}  
		\caption{$\gamma$-SOM, Dijkstra's path planning}
		\label{fig:gammaSOM_dijkstra}
	\end{subfigure}
	\begin{subfigure}{.49\textwidth}
	    \centering
	    \includegraphics[height=3cm]{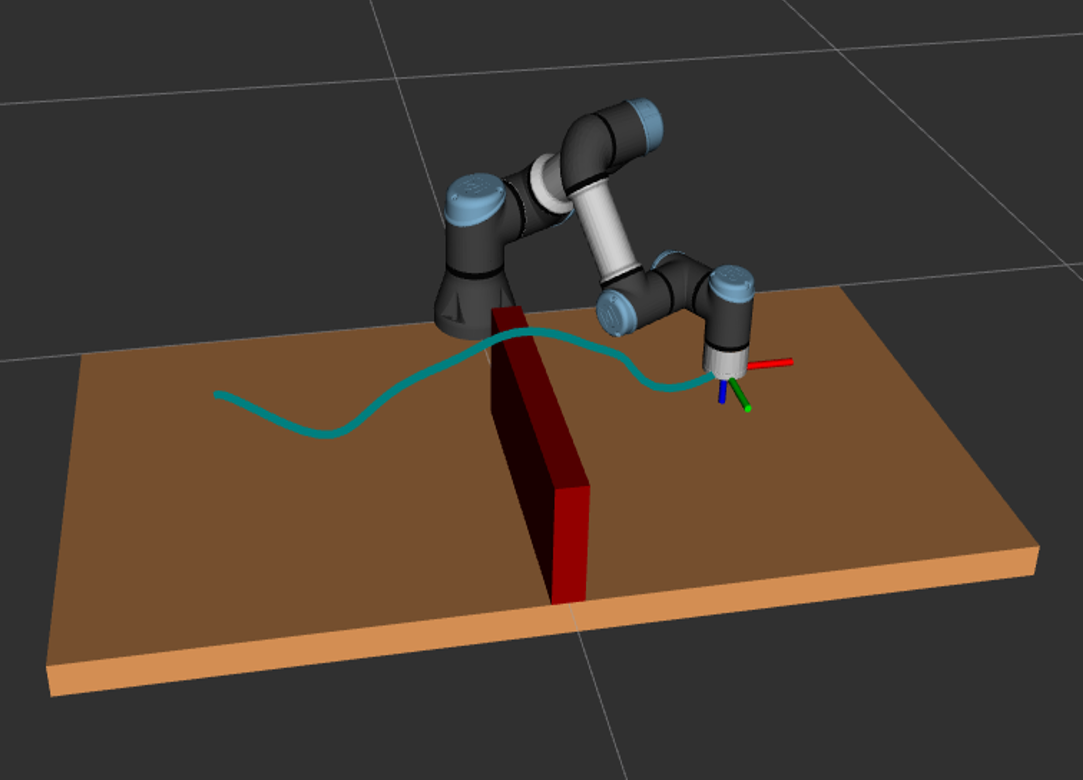}  
	    \hspace{0.35cm}
	    \includegraphics[height=3cm]{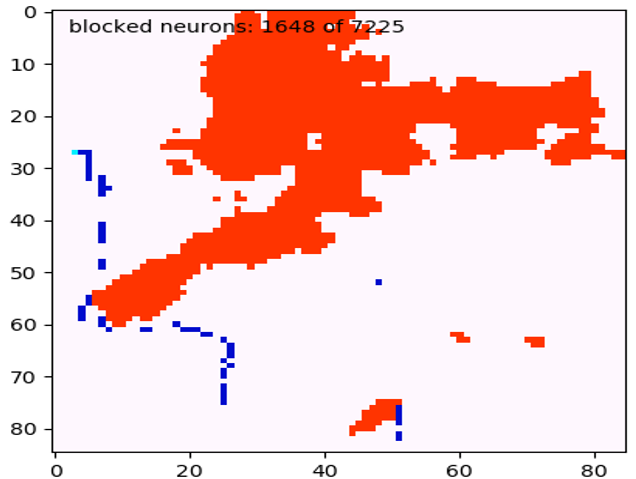}  
	    \caption{$\gamma$-SOM, Dijkstra's, smoothed path}
	    \label{fig:gammaSOM_dijkstra_smoothed}
    \end{subfigure}
    \begin{subfigure}{.49\textwidth}
		\centering
		\includegraphics[height=3cm]{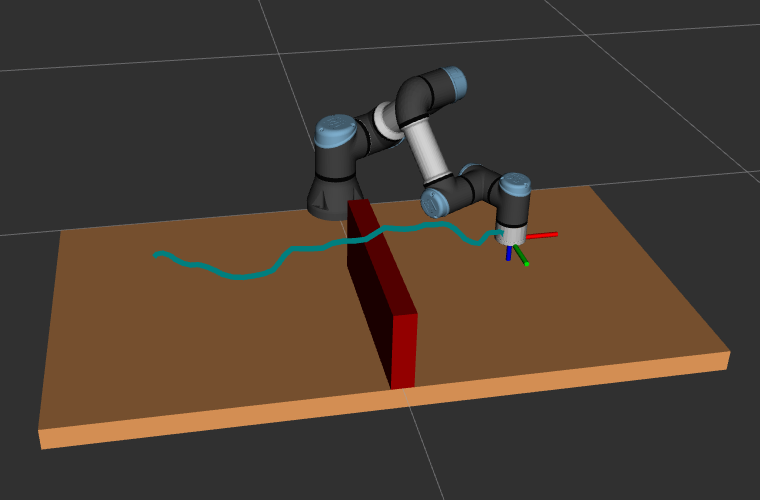}  
        \hspace{0.35cm}
		\includegraphics[height=3cm]{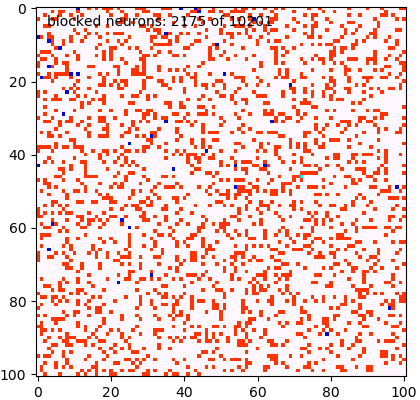}  
		\caption{GNG, Dijkstra's path planning}
		\label{fig:GNG_dijkstra}
	\end{subfigure}
	\begin{subfigure}{.49\textwidth}
    	\centering
    	\includegraphics[height=3cm]{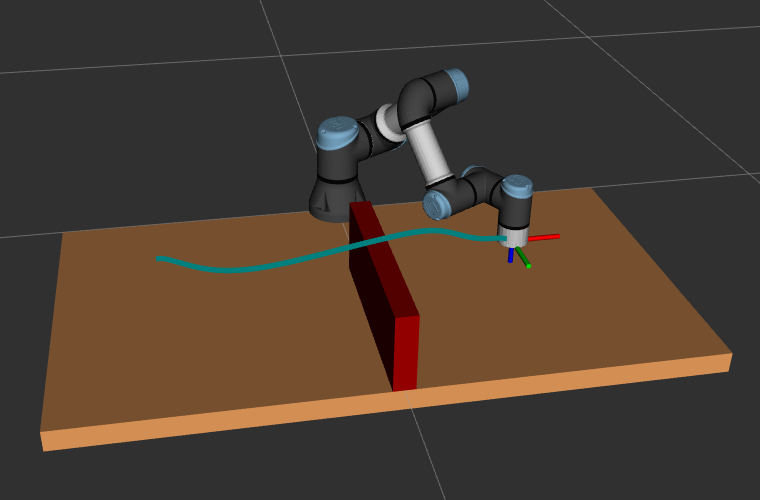}  
       \hspace{0.35cm}
    	\includegraphics[height=3cm]{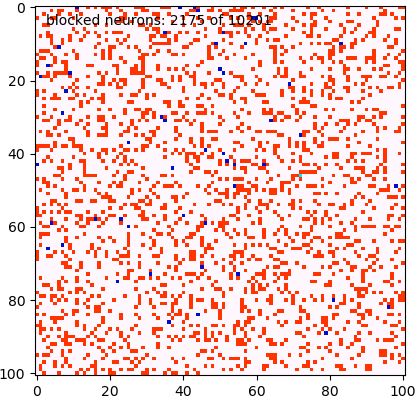}  
    	\caption{GNG, Dijkstra's, smoothed path}
    	\label{fig:GNG_dijkstra_smoothed}
    \end{subfigure}
	\caption{Planned paths regarding the end-effector in the task space on the left and the respective cognitive maps on the right. In the reduced C-space red represents the obstacles and blue the path. While obstacles and the planned path are visible in the cognitive map of the $\gamma$-SOM in \protect\subref{fig:gammaSOM_wavefront},~\protect\subref{fig:gammaSOM_dijkstra}~\&~\protect\subref{fig:gammaSOM_dijkstra_smoothed} the GNG's output space in \protect\subref{fig:GNG_dijkstra}~\&~\protect\subref{fig:GNG_dijkstra_smoothed} is scattered. However, the generated trajectories are purposeful and direct for the GNG and more twisted for the $\gamma$-SOM. 
	By comparing \protect\subref{fig:gammaSOM_wavefront}~\&~\protect\subref{fig:gammaSOM_dijkstra} it can be seen that the Wavefront algorithm finds the shortest path within the cognitive
	map while Dijkstra’s algorithm produces more optimal trajectories in the task space.
	Planning times are stated in~\autoref{tab:comparison_sonn_versions}. }
		
	\label{fig:sim_and_maps}
\end{figure}

A comparison of the planned paths by the Wavefront and Dijkstra’s algorithm is shown in~\autoref{fig:sim_and_maps} \subref{fig:gammaSOM_wavefront} \& \subref{fig:gammaSOM_dijkstra}. On the right side, the output space of the $\gamma$-SOM is depicted, which represents the cognitive map of the robot. It can be observed that both, the Wavefront and Dijkstra's algorithm, plan a path around the blocked neurons in red. While the Wavefront algorithm finds the visually shortest path on the cognitive map with 72 neurons on it (see \autoref{tab:comparison_sonn_versions}), the path of Dijkstra's algorithm seems to take a detour with 76 neurons. However, the trajectories which the robot performs are more optimal for Dijkstra's path, as illustrated in \autoref{fig:gammaSOM_wavefront}~\&~\subref{fig:gammaSOM_dijkstra} respectively. The detours in the cognitive map visualization origin from the fact, that the distance of two adjacent neurons in the SOM structure is not evenly distributed, but varies. Thus, the shortest distanced path found by Dijkstra's algorithm differs from the Wavefront's path with the least neurons. 
The observed behavior of the two algorithms is consistent with our expectation from \autoref{sec:cognitive_map}. As the learned weights of the neurons are not evenly distributed, the generated space corresponds to a weighted graph. The complete and optimal grid search of the Wavefront neglects weighted edges. Hence, Dijkstra’s algorithm, considering weighted edges, generates more optimal paths in the reduced C-space.
The same effect applies to GNGs.
Although Dijkstra's algorithm is in theory computationally more expensive, the experiments showed no measurable difference for execution times as stated in \autoref{tab:comparison_sonn_versions}.
\begin{table}[h!]
	\centering
	\begin{tabular}{@{}lccccc}
		\toprule
		SONN    		& algorithm 	& s   	& \#N 	& $\mu$ 			& $\sigma$ \\ \midrule
		$\gamma$-SOM 	& Wavefront 	& no    & 72 	& 0.007 (0.006) 	& 0.0013 \\
		$\gamma$-SOM 	& Dijkstra's 	& no 	& 76 	& 0.007 (0.006) 	& 0.0004 \\
		$\gamma$-SOM 	& Dijkstra's 	& yes  	& 76 	& 0.088 (0.085)  	& 0.0028 \\ 
		GNG 			& Dijkstra's 	& no  	& 39 	& 0.013 (0.013) 	& 0.0006 \\ 
		GNG 			& Dijkstra's 	& yes  	& 39 	& 0.037 (0.037)  	& 0.0012 \\ 
		\bottomrule
	\end{tabular}
	\caption{Number of neurons (\#N) within a path and planning times for different models and algorithms. The column \textit{s} indicates whether it was smoothed. The mean $\mu$ and standard deviation $\sigma$ are given for 20 planning attempts in seconds. The numbers in brackets refer to the runs in~\autoref{fig:sim_and_maps}.}
	\label{tab:comparison_sonn_versions}
\end{table}

\subsection{Obstacle avoidance with $\gamma$-SOM and GNG} \label{sec:som_vs_gng}
The comparison between the $\gamma$-SOM (\autoref{fig:gammaSOM_dijkstra}) and the GNG (\autoref{fig:GNG_dijkstra}) shows the general difference in their cognitive map structure, that is their output space.
While the topology of the $\gamma$-SOM lies in its rigid 2D-structure which connects two neighbored neurons, the topology of the GNG output space is based on the learned synapses, connecting neurons all over the map.
Thus, the blocked neuron regions and the planned path in the cognitive map of the GNG is not as continuous as for the $\gamma$-SOM but scattered. 
However, for the $\gamma$-SOM the generated motions during the trajectory execution appear to be more twisted than for the GNG.
This can be explained by the 2D rigid structure of the SOM based model which exhibits a topological mismatch with the multi-dimensional C-space. In contrast, the GNG learns the connections explicitly and can match every topological dimensionality~\cite{Steffen2022_dimreduction2}.
To smooth the trajectories, non-uniform rational B-spline (NURBS) curves were used~\cite{Ravankar2018}. The smoothed paths in the cognitive maps are shown in \autoref{fig:gammaSOM_dijkstra_smoothed}~\&~\subref{fig:GNG_dijkstra_smoothed} for $\gamma$-SOM and GNG respectively. 
Note that the smoothed path shows gaps in the cognitive map of the $\gamma$-SOM and other activated neurons not directly neighbored to the original path. Apparently, trajectory smoothing results in the omission of certain neurons on the path that are responsible for the strongest trajectory twists. Also smoothed configurations may correspond to neurons not directly neighbored in the cognitive map. Eventually, we receive smooth motions around the obstacle. 
In \autoref{tab:comparison_sonn_versions} can be seen, that the $\gamma$-SOM holds more neurons on the path, which means a higher path resolution. 
Higher path resolutions are generally desirable since the sections in between two configurations on a path cannot be guaranteed to be collision-free if the distances are larger than the smallest obstacles.
However, the $\gamma$-SOM often failed to find a path for bigger obstacles because too many neurons were blocked by the obstacles. In contrast, the GNG showed a better generalization and was also able to find paths around obstacles which blocked large parts of the robots task space.
Regarding the execution times, the actual planning is slightly faster for the $\gamma$-SOM than for the GNG. The reason are less synapses between the neurons of the $\gamma$-SOM. 
The computation time for the smoothing is higher for the $\gamma$-SOM due to more neurons on the path.
However, the planning times with less than 0.02 s are remarkable fast and even inclusive smoothing which seems to be the most costly operation in the planning pipeline, the execution times are less than 0.1 second, which emphasize the real-time capability of the presented approach.
It is also noteworthy, that due to the growing learning behavior, the GNG can easily be extended by additional training trajectories. This means, a pre-trained GNG network covering the general work space of the robot could be used as basis and specific task trajectories could be additionally trained subsequently. This allows firstly a shortened training time of the network and secondly results in different areas within the network for different tasks, similar different specified regions which can be found in the brain. 
In contrast, the $\gamma$-SOM cannot be extended but must be retrained, due to its rigid architecture.

\subsection{Comparisons with modern sample-based planners in ROS} \label{sec:comparison_sample_based}
\begin{figure*}
	\centering
	\begin{subfigure}{.3\textwidth}
		\centering
		\includegraphics[height=3.5cm]{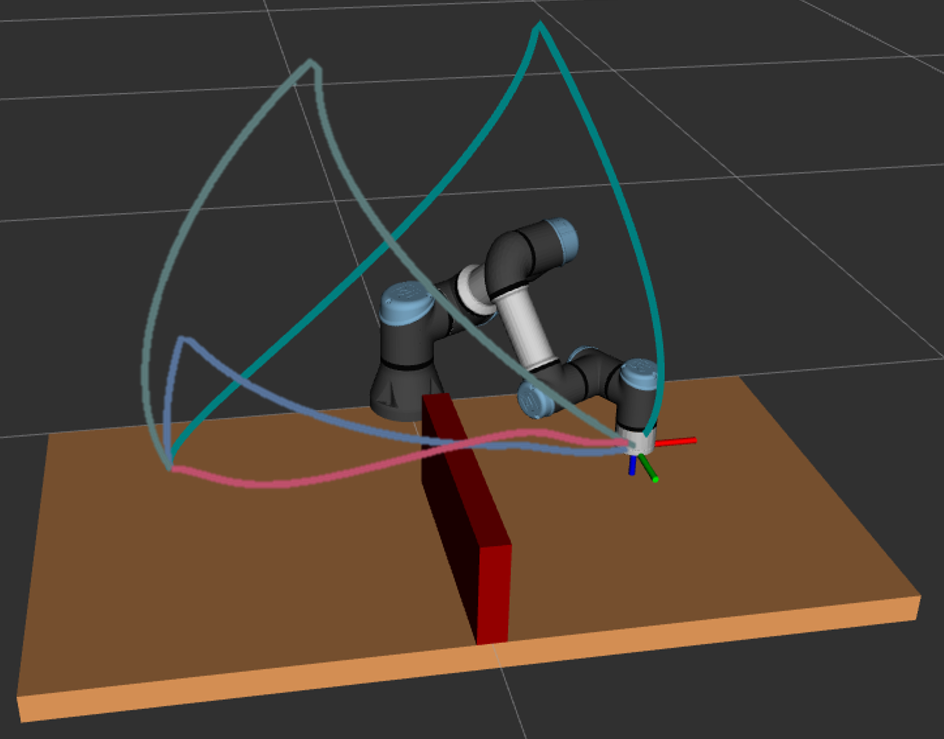}  
		\caption{PRM}
		\label{fig:PRM}
	\end{subfigure}
	\begin{subfigure}{.3\textwidth}
		\centering
		\includegraphics[height=3.5cm]{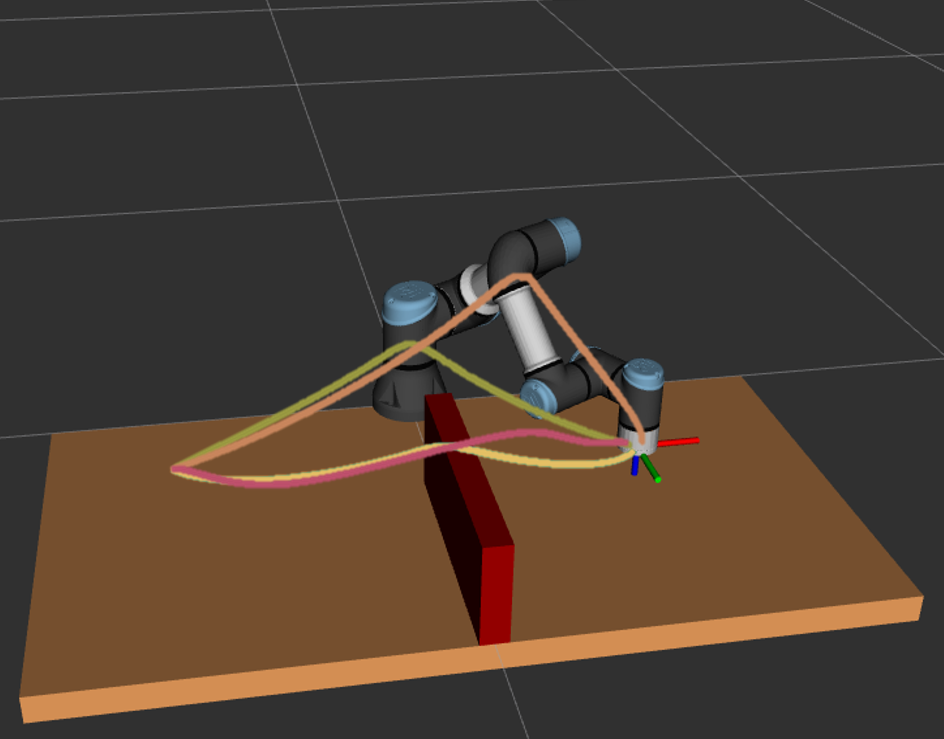}  
		\caption{RRT}
		\label{fig:RRT}
	\end{subfigure}
	\begin{subfigure}{.3\textwidth}
		\centering
		\includegraphics[height=3.5cm]{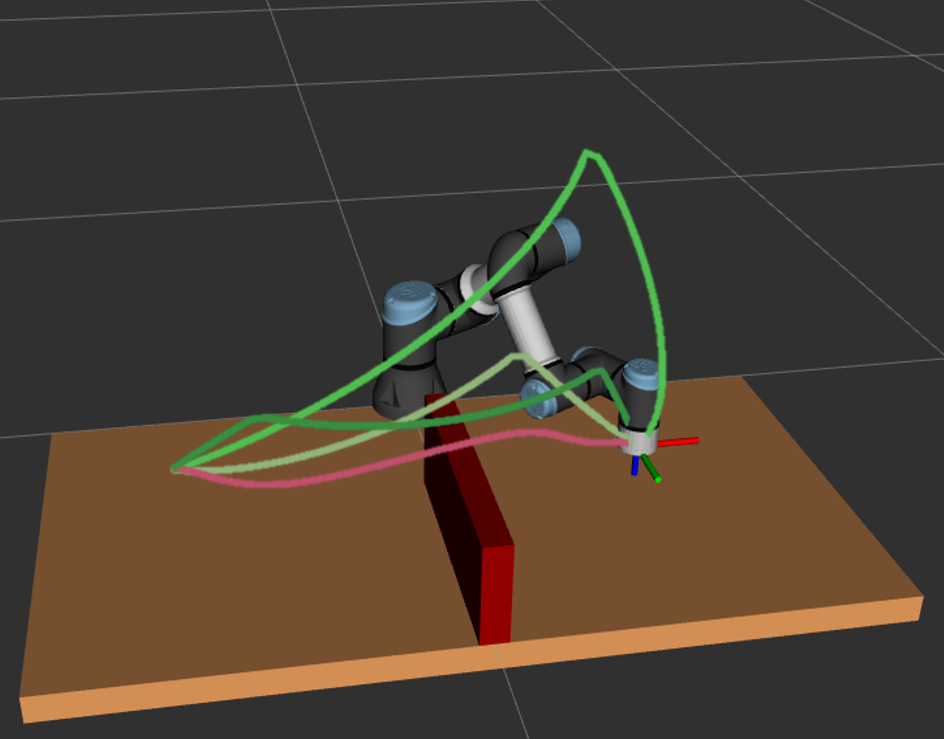}  
		\caption{RRTConnect}
		\label{fig:RRTConnect}
	\end{subfigure}
	\caption{Trajectories of probabilistic sample-based planners compared to the smoothed GNG curve from~\autoref{fig:GNG_dijkstra_smoothed}, displayed in red. For each sample-based planner, several different trajectories are displayed to illustrate their probabilistic character.}
	\label{fig:conventional_planners}
\end{figure*}
The proposed approach achieves the best results with Dijkstra's algorithm in the reduced space of the GNG as shown in~\autoref{fig:GNG_dijkstra}~\&~\subref{fig:GNG_dijkstra_smoothed}. To provide a comparison regarding common sample-based planners, MoveIt\footnote{\url{https://moveit.ros.org/}} was used.
Comparative trajectories created with the PRM, RRT and RRTConnect not executed in the reduced space are shown in \autoref{fig:conventional_planners}. 
For every planner, several different trajectories are displayed in different
colors to illustrate the probabilistic character of the RRT and PRM. To enable a simple comparison, all motions had the same start and target configuration as in~\autoref{fig:sim_and_maps}. Also the generated paths in \autoref{fig:GNG_dijkstra_smoothed}, the smoothed trajectory of Dijkstra's algorithm in the reduced space of the GNG, is drawn in again here. 
None of the sample-based planner's paths is as optimal as the best path of the proposed approach. 

\begin{table}[h!]
	\centering
	\begin{tabular}{lcc}
		\toprule
		Planner & $\mu$  & $\sigma$  \\ 
		\midrule
		GNG (Dijkstra's + path smoothing)	& 0.037 & 0.0012\\ 
		PRM        							& 0.115 & 0.1161 \\
		RRT        							& 2.876 & 1.0126 \\
		RRTConnect 							& 0.023 & 0.0045 \\ 
		\bottomrule         
	\end{tabular}
	\caption{Results of the proposed approach compared to several sample-based planners. Mean $\mu$ and standard deviation $\sigma$ for 20 runs. Planning time is in seconds.
	\label{tab:conventional_planners}}
\end{table}

\iffalse
\begin{table}[h!]
	\centering
	\begin{tabular}{lr@{}}
		\toprule
		Planner & \multicolumn{1}{l}{Plan Time (s)} \\ 
	 	\midrule
	 	GNG \& Dijkstra's 	& 0.037 \\ 
	 	PRM        			& 0.097 \\
		            		& 0.069 \\
		            		& 0.077 \\
		RRT        			& 2.167 \\
							& 1.354 \\ 
							& 1.138 \\
		RRTConnect 			& 0.025 \\
		           			& 0.025 \\
		         			& 0.023 \\ 
	   \bottomrule         
	\end{tabular}
	\caption{Planning times of the trajectories in \autoref{fig:conventional_planners} with the sample-based planners PRM, RRT and RRTConnect in MoveIt.}
	\label{tab:conventional_planners}
\end{table}
\fi
Although two of the three trajectories generated with the RRTConnect are quite close to the generated path of the presented method. 
Even though, an optimal planner is used for the proposed approach, the planning time inclusive smoothing is than for the common sample-based planners RRT and PRM and shows similar planning times as the advanced RRTConnect as stated in~\autoref{tab:conventional_planners}. 
Furthermore, the sample-based planners generally have difficulties with many obstacles~\cite{Kiesel2017} and narrow passages~\cite{Hsu2003, Szkandera2020, Orthey2021} since these are rather difficult to sample. 
In contrast, the presented approach is independent of random sampling and should not be affected from multiple obstacles or narrow passages as long as any paths were learned in these regions.
Furthermore, constrained motions can be taught and subsequently planned and executed. If the SONN is trained with data in which the wrist is kept constant for example, all generated motions keep the orientation of the wrist link constant. Considering such constraints is far from trivial for common planners and needs additional computational effort~\cite{Milton1994, Sucan2012}.

\section{CONCLUSIONS} \label{sec:conclusions}
A biologically inspired highly-reactive, collision-free path planner was designed and implemented on a 6~DOF robot. At the core of the presented concept is a SONN. Its output space approximates the reduced C-space of the robot and can be interpreted as a neural cognitive map in which planning takes place. Due to the reduced search space, optimal planning algorithms can be used. Thereby, we could improve the trajectory quality using Dijkstra's algorithm instead of the Wavefront algorithm from former works. To allow fast obstacle avoidance, a lookup table is implemented that associates occupied cells of the task space with all affected joint configurations. Hence, neurons which would lead to a collision are blocked. 
As is the case with many learning methods, the quality of the proposed concept is only as good as the data used to train the SONN. Hereby, the training data and the use case must be compatible to guarantee a high accuracy of planned trajectories. 
Also, the path resolution depends on the ratio between the number of training trajectories and the number of neurons. The more neurons learn a single trajectory, the higher the path resolution. However, a larger network increases the amount of memory needed for the SONN and especially for the lookup table. A high path resolution is an important factor regarding the obstacle avoidance accuracy. Sparser sampled paths mean larger joint angle jumps between the individual planned joint states along the trajectory. If jumps between two states are larger than a minimal obstacle size, there is no guarantee that the interpolated trajectory is collision-free. Thus, a good balance between path resolution and memory requirements is crucial.
It is also noteworthy that in the approach, there is no distance measure between the obstacles and the robot. 
Instead, collisions are avoided by repeatedly planning and updating the shortest collision-free path in the reduced C-space. This is very fast and efficient, but the consideration of distances in the task space is complicated.
Hence, distance related tasks like backing off or keeping a precise distance, are not feasible.
The evaluation in a robot work cell with a vision component demonstrated the applicability of the approach for high-reactive and collision-free path planning. The robot reacts to dynamically changing environments and re-plans its trajectories within 0.02 seconds (see \autoref{tab:comparison_sonn_versions}).
This proofs the real-time capability of our concept and shows significantly faster planning times than RRT or PRM and is comparable to RRTConnect.
Even including path smoothing, the pipeline's computationally most intensive part, planning times stay below 0.1 seconds.
The evaluation of two different SONN versions showed that the $\gamma$-SOM generates coherent paths in the cognitive map. However, in terms of performance and path quality in the work space, the GNG is far superior. In contrast to traditional probabilistic sample-based planners like RRT and PRM, the paths generated by our system showed to be reproducible and optimal within the learned C-space representation, regarding the path length. 
Traditional planners, struggle with cluttered scenes and many DOF, as they have difficulties to sample the free C-space efficiently. In contrast, the presented approach does not randomly sample the whole C-space but learns the effectively used sub space. 
Furthermore, configuration constraints are easy to implement with this method. If the SONN is trained exclusively with data that keeps the desired joint angle constant, all generated motions
keep the orientation as well. Also, the GNG structure allows easy extensions with additional training data. Thus, a pre-trained general network, could be extended to a specific task by manual hand guidance. \\
A naive improvement of this method is to increase the SONN size for a higher path resolution. To overcome the issues regarding memory requirements for lookup tables, bit encoding~\cite{Wu2005} could be used.
In this work, weight vectors of the SONNs consisted solely of joint angle configurations. However, they could be extended to velocities, accelerations or torques to enable path planning with respect to additional optimality criteria.
Planning times could be improved using the target BMU as well as the start BMU as seed points for the graph search, similar to the RRTConnect \cite{Kuffner2000} which showed substantially faster planning times than the standard RRT. As smoothing is the computationally heaviest burden of the proposed planning pipeline, faster smoothing techniques could be evaluated. Additionally, the concept could be transferred to Spiking Neural Networks (SNN) and neuromorphic hardware to take full advantage of SNN’s high parallelization potential. 
Finally, it would be very interesting to evaluate the method in cluttered scenes or narrow passages. The presented approach is independent of random sampling and should, in theory, not be affected in these complex scenes as much as traditional sample-based planners.

\section*{ACKNOWLEDGMENT}
This research has been supported by the European Union's Horizon 2020 Framework Programme for Research and Innovation under the Specific Grant Agreement No. 945539 (Human Brain Project SGA3).

\bibliographystyle{IEEEtran}
\bibliography{bib}

\begin{thebibliography}{10}
\providecommand{\url}[1]{#1}
\csname url@rmstyle\endcsname
\providecommand{\newblock}{\relax}
\providecommand{\bibinfo}[2]{#2}
\providecommand\BIBentrySTDinterwordspacing{\spaceskip=0pt\relax}
\providecommand\BIBentryALTinterwordstretchfactor{4}
\providecommand\BIBentryALTinterwordspacing{\spaceskip=\fontdimen2\font plus
\BIBentryALTinterwordstretchfactor\fontdimen3\font minus
  \fontdimen4\font\relax}
\providecommand\BIBforeignlanguage[2]{{%
\expandafter\ifx\csname l@#1\endcsname\relax
\typeout{** WARNING: IEEEtran.bst: No hyphenation pattern has been}%
\typeout{** loaded for the language `#1'. Using the pattern for}%
\typeout{** the default language instead.}%
\else
\language=\csname l@#1\endcsname
\fi
#2}}

\bibitem{Liu2021}
Y.~Liu, F.~Zha, M.~Li, W.~Guo, Y.~Jia, P.~Wang, Y.~Zang, and L.~Sun,
  ``{Creating Better Collision-Free Trajectory for Robot Motion Planning by
  Linearly Constrained Quadratic Programming},'' \emph{Frontiers in
  Neurorobotics}, vol.~15, p. 104, 2021.

\bibitem{Kiesel2017}
S.~Kiesel, T.~Gu, and W.~Ruml, ``{An effort bias for sampling-based motion
  planning},'' \emph{IROS}, pp. 2864--2871, 2017.

\bibitem{Hsu2003}
D.~Hsu, T.~Jiang, J.~Reif, and Z.~Sun, ``{The bridge test for sampling narrow
  passages with probabilistic roadmap planners},'' \emph{ICRA}, vol.~3, pp.
  4420--4426, 2003.

\bibitem{Szkandera2020}
J.~Szkandera, I.~Kolingerov{\'{a}}, and M.~Maň{\'{a}}k, ``{Narrow Passage
  Problem Solution for Motion Planning},'' \emph{Lect. Notes Comput. Sci.},
  vol. 12137, pp. 459--470, 2020.

\bibitem{Orthey2021}
A.~Orthey and M.~Toussaint, ``{Section Patterns: Efficiently Solving Narrow
  Passage Problems in Multilevel Motion Planning},'' \emph{IEEE Trans. on
  Robotics}, 2021.

\bibitem{Wise2001}
K.~D. Wise and A.~Bowyer, ``{A survey of global configuration-space mapping
  techniques for a single robot in a static environment},'' \emph{Int. J. of
  Robotics Research}, vol.~19, no.~8, pp. 762--779, 2001.

\bibitem{Brost1989}
R.~Brost, ``{Computing metric and topological properties of configuration-space
  obstacles.}'' \emph{ICRA}, pp. 170--176, 1989.

\bibitem{Varadhan2006}
G.~Varadhan, Y.~J. Kim, S.~Krishnan, and D.~Manocha, ``{Topology preserving
  approximation of free configuration space},'' \emph{ICRA}, pp. 3041--3048,
  2006.

\bibitem{DAvella2003}
A.~D'Avella, P.~Saltiel, and E.~Bizzi, ``{Combinations of muscle synergies in
  the construction of a natural motor behavior},'' \emph{Nature Neuroscience},
  vol.~6, no.~3, pp. 300--308, 2003.

\bibitem{Pan2015}
J.~Pan and D.~Manocha, ``{Efficient Configuration Space Construction and
  Optimization for Motion Planning},'' \emph{Engineering}, vol.~1, pp.
  046--057, 2015.

\bibitem{Ward2007}
J.~Ward and J.~Katupitiya, ``{Free space mapping and motion planning in
  configuration space for mobile manipulators},'' \emph{ICRA}, pp. 4981--4986,
  2007.

\bibitem{Han2021}
B.~Han, X.~Luo, Q.~Luo, Y.~Zhao, and B.~Lin, ``{Research on Obstacle Avoidance
  Motion Planning Technology of 6-DOF Manipulator},'' \emph{Adv. Intell. Syst.
  Comput.}, vol. 1296, pp. 604--614, 2021.

\bibitem{Wu2005}
X.~Wu, Q.~Lit, and K.~H. Heng, ``{A new algorithm for construction of
  discretized configuration space obstacle and collision detection of
  manipulators},'' \emph{ICAR}, pp. 90--95, 2005.

\bibitem{Xie2020}
Y.~Xie, R.~Zhou, and Y.~Yang, ``{Improved distorted configuration space path
  planning and its application to robot manipulators},'' \emph{Sensors},
  vol.~20, no.~21, pp. 1--23, 2020.

\bibitem{Huerta-Chua2021}
J.~Huerta-Chua, G.~Diaz-Arango, H.~Vazquez-Leal, J.~Flores-Mendez,
  M.~Moreno-Moreno, R.~C. Ambrosio-Lazaro, and C.~Hernandez-Mejia, ``{Exploring
  a Novel Multiple-Query Resistive Grid-Based Planning Method Applied to
  High-DOF Robotic Manipulators},'' \emph{Sensors}, vol.~21, no.~9, 2021.

\bibitem{Miljkovic2017}
D.~Miljkovic, ``{Brief review of self-organizing maps},'' in \emph{MIPRO},
  2017, pp. 1061--1066.

\bibitem{VanHulle2012}
M.~M. {Van Hulle}, ``{Self-Organizing Maps},'' Tech. Rep., 2012.

\bibitem{Steffen2021_dimreduction}
L.~Steffen, K.~Glueck, S.~Ulbrich, A.~Roennau, and D.~Dillmann, ``{Reducing the
  Dimension of the Configuration Space with Self Organizing Neural Networks},''
  \emph{ICARM}, 2021.

\bibitem{Steffen2022_dimreduction2}
L.~Steffen, T.~Weyer, K.~Glueck, S.~Ulbrich, A.~Roennau, and R.~Dillmann, ``{A
  Comparison of Self-Organizing Neural Networks to Reduce the High
  Dimensionality of the Configuration Space of Robots},'' \emph{Submitted for
  publication.}, vol.~17, 2022.

\bibitem{Kuffner2000}
J.~J. Kuffner and S.~M. {La Valle}, ``{RRT-connect: an efficient approach to
  single-query path planning},'' \emph{ICRA}, vol.~2, pp. 995--1001, 2000.

\bibitem{Sucan2012}
I.~A. Şucan, M.~Moll, and L.~Kavraki, ``{The open motion planning library},''
  \emph{IEEE Robot Autom Mag}, vol.~19, no.~4, pp. 72--82, 2012.

\bibitem{Kohonen1982}
T.~Kohonen, ``{Self-organized formation of topologically correct feature
  maps},'' \emph{Biological Cybernetics}, vol.~43, no.~1, pp. 59--69, jan 1982.

\bibitem{Fritzke1995}
B.~Fritzke, ``{A Growing Neural Gas Network Learns Topologies},'' \emph{Adv. in
  Neural Information Processing Systems}, 1995.

\bibitem{Estevez2009}
P.~A. Est{\'{e}}vez and R.~Hern{\'{a}}ndez, ``{Gamma SOM for temporal sequence
  processing},'' \emph{Lect. Notes Comput. Sci.}, vol. 5629, pp. 63--71, 2009.

\bibitem{Tolman1948}
E.~C. Tolman, ``{Cognitive maps in rats and men.}'' \emph{Psychological
  Review}, vol.~55, no.~4, pp. 189--208, 1948.

\bibitem{OKeefe1971}
J.~O'Keefe and J.~Dostrovsky, ``{The hippocampus as a spatial map. Preliminary
  evidence from unit activity in the freely-moving rat},'' \emph{Brain
  Research}, vol.~34, no.~1, pp. 171--175, 1971.

\bibitem{OKeefe1978}
J.~O'Keefe and L.~Nadel, \emph{{The hippocampus as a cognitive map}}.\hskip 1em
  plus 0.5em minus 0.4em\relax Clarendon Press, 1978.

\bibitem{Derdikman2010}
D.~Derdikman and E.~I. Moser, ``{A manifold of spatial maps in the brain},''
  \emph{Trends Cogn. Sci.}, vol.~14, no.~12, pp. 561--569, 2010.

\bibitem{Huang2009}
C.~Y. Huang, C.~Y. Lai, and K.~T. Cheng, ``{Fundamentals of Algorithms},''
  \emph{Electronic Design Automation}, pp. 173--234, 2009.

\bibitem{Hermann2014}
A.~Hermann, F.~Drews, J.~Bauer, S.~Klemm, A.~Roennau, and R.~Dillmann,
  ``{Unified GPU voxel collision detection for mobile manipulation planning},''
  in \emph{IROS}, 2014, pp. 4154--4160.

\bibitem{Quigley2009}
M.~Quigley, B.~Gerkey, K.~Conley, J.~Faust, T.~Foote, J.~Leibs, E.~Berger,
  R.~Wheeler, and A.~Ng, ``{ROS: An open-source Robot Operating System},''
  \emph{ICRA}, 2009.

\bibitem{Thomas2014}
D.~Thomas, W.~Woodall, and E.~Fernandez, ``{Next-generation ROS: Building on
  DDS},'' \emph{ROSCon}, 2014.

\bibitem{Ravankar2018}
A.~Ravankar, A.~A. Ravankar, Y.~Kobayashi, Y.~Hoshino, and C.~C. Peng, ``{Path
  smoothing techniques in robot navigation: State-of-the-art, current and
  future challenges},'' \emph{Sensors}, vol.~18, no.~9, 2018.

\bibitem{Milton1994}
C.~Milton and D.~Jennings, ``{Path planning for a manipulator with constraints
  on orientation},'' \emph{IROS}, vol.~11, no. 1-2, pp. 67--77, 1994.

\end{thebibliography}

\end{document}